\documentclass{article}

\usepackage[utf8]{inputenc}
\usepackage{arxiv}
\usepackage{graphicx}
\usepackage{subcaption}
\usepackage{booktabs}
\usepackage{multirow}
\usepackage{pifont}
\usepackage{adjustbox}

\usepackage{tikz}
\usetikzlibrary{positioning, arrows.meta, shapes.geometric, calc}
\usepackage{pgfplots}
\pgfplotsset{width=10cm,compat=1.9}

\usepackage{xcolor}
\usepackage{hyperref}
\usepackage{tabularx}
\usepackage{booktabs}
\usepackage{amsmath}
\usepackage{cleveref}
\usepackage{caption}
\usepackage{graphicx}
\usepackage{float}
\usepackage{booktabs}
\usepackage{tabularx}
\usepackage{caption}
\usepackage{amsmath}
\usepackage{microtype}
\usepackage{authblk}
\usepackage{adjustbox}

\title{Gesture Classification in Artworks Using Contextual Image Features}

\newbox{\orcid}\sbox{\orcid}{\includegraphics[scale=0.06]{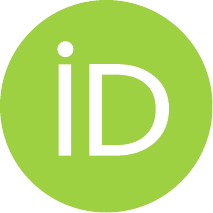}} 
\author[1]{%
	\href{https://orcid.org/0009-0008-8125-7081}{\usebox{\orcid}\hspace{1mm}Azhar Hussian\thanks{\texttt{azhar.hussian@fau.de}}}%
}
\author[1]{%
	\href{https://orcid.org/0000-0003-4366-5216}{\usebox{\orcid}\hspace{1mm}Mathias Zinnen}%
}
\author[1]{%
	\href{https://orcid.org/0009-0004-0812-9316}{\usebox{\orcid}\hspace{1mm}Thi My Hang Tran}%
}
\author[1]{%
	\href{https://orcid.org/0000-0002-9550-5284}{\usebox{\orcid}\hspace{1mm}Andreas Maier}%
}
\author[1]{%
	\href{https://orcid.org/0000-0003-0455-3799}{\usebox{\orcid}\hspace{1mm}Vincent Christlein}%
}
\affil[1]{Pattern Recognition Lab, Friedrich-Alexander-Universität, Erlangen, Germany}

\DeclareUnicodeCharacter{0308}{\textbf{{\"i}}}

\begin{document}

\maketitle

\begin{abstract}
Recognizing gestures in artworks can add a valuable dimension to art understanding and help to acknowledge the role of the sense of smell in cultural heritage. We propose a method to recognize smell gestures in historical artworks. We show that combining local features with global image context improves classification performance notably on different backbones.
\end{abstract}

\begin{figure*}[!h]
    \centering
    \begin{subfigure}[t]{0.15\textwidth}
        \adjustbox{valign=t}{\includegraphics[width=\linewidth, height=\linewidth]{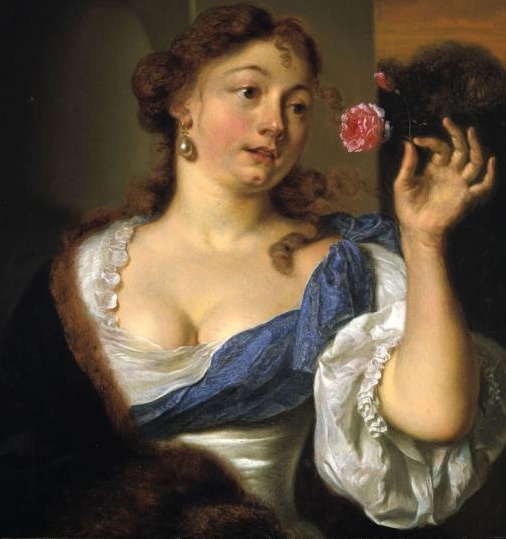}}
        \caption{Sniffing}
        \label{fig:sub1}
    \end{subfigure}
    \hfill
    \begin{subfigure}[t]{0.15\textwidth}
        \adjustbox{valign=t}{\includegraphics[width=\linewidth, height=\linewidth]{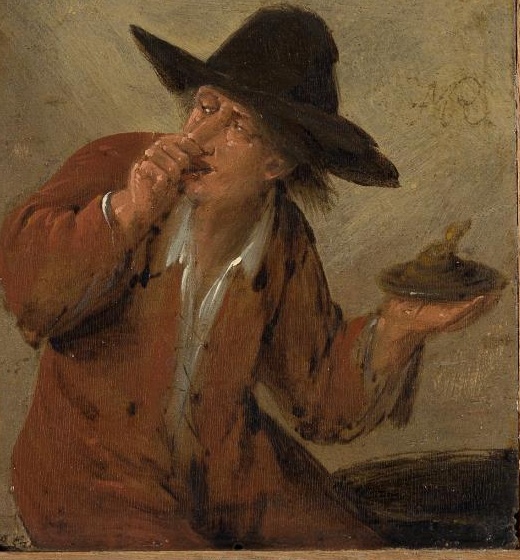}}
        \caption{Holding the nose}
        \label{fig:sub2}
    \end{subfigure}
    \hfill
    \begin{subfigure}[t]{0.15\textwidth}
        \adjustbox{valign=t}{\includegraphics[width=\linewidth, height=\linewidth]{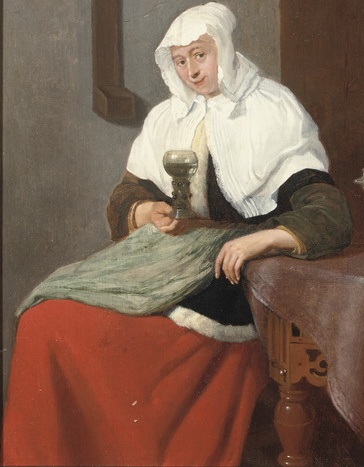}}
        \caption{Drinking}
        \label{fig:sub4}
    \end{subfigure}
    \hfill
    \begin{subfigure}[t]{0.15\textwidth}
        \adjustbox{valign=t}{\includegraphics[width=\linewidth, height=\linewidth]{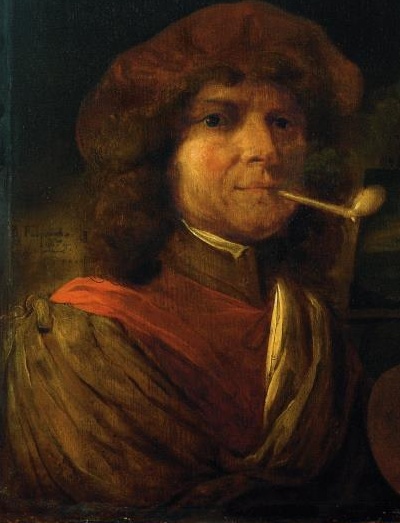}}
        \caption{Smoking}
        \label{fig:sub5}
    \end{subfigure}
    \hfill
    \begin{subfigure}[t]{0.15\textwidth}
        \adjustbox{valign=t}{\includegraphics[width=\linewidth, height=\linewidth]{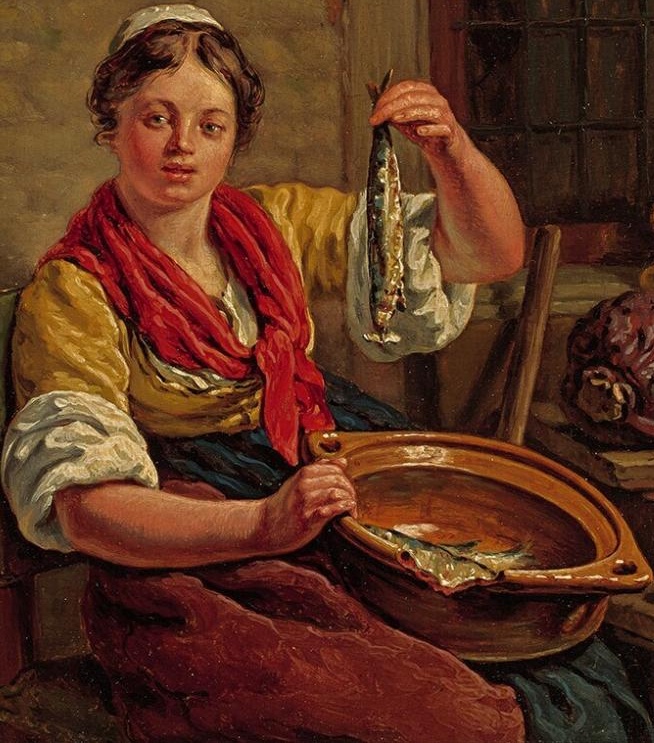}}
        \caption{Cooking}
        \label{fig:sub6}
    \end{subfigure}
    \hfill
    \begin{subfigure}[t]{0.15\textwidth}
        \adjustbox{valign=t}{\includegraphics[width=\linewidth, height=\linewidth]{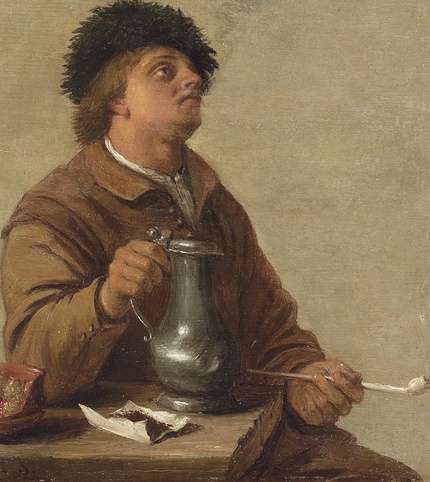}}
        \caption{c and d}
        \label{fig:sub3}
    \end{subfigure}
    \caption{Example of each class in SniffyArt Dataset~\cite{zinnen2023sniffyart}}
    \label{fig:classes_example}
\end{figure*}

\section{Introduction}

Smell gestures can provide a valuable gateway to the history of olfaction. 
We propose a new method to recognize smell gestures in historical artworks under the challenging conditions of a low-data regime and a large class imbalance and show that combining local features with global image context improves classification accuracy. We use the SniffyArt~\cite{zinnen2023sniffyart} dataset that has been introduced to facilitate research in the classification of gestures within artwork images.

We propose a two-step smell gesture recognition method that first detects persons depicted in the artwork and then classifies them according to one of six smell gestures. Well-established object detection approaches can be used for the person detection step.
With the availability of large-scale image datasets such as COCO~\cite{lin2015microsoft} or  OpenImages~\cite{kuznetsova2020open}, object detection algorithms like the canonical Faster-R-CNN~\cite{ren2016faster}, YOLO~\cite{redmon2016you} or the more recent DETR-based~\cite{carion2020end} algorithms such as DINO~\cite{zhang2022dino} have shown impressive performance in detecting a large number of categories in natural images. However, their performance has been shown to drop significantly when applied to artistic data~\cite{westlake2016detecting,reshetnikov2022deart,zinnen2022odor}.

In the realm of classification, traditional Convolutional Neural Network (CNN) architectures like ResNet~\cite{he2016deep} and EfficientNet~\cite{tan2019efficientnet} have been widely adopted because of their success on the ImageNet~\cite{russakovsky2015imagenet} dataset. Similar to object detection, transformers have gained widespread adoption for classification tasks. The increasing popularity of attention-based models such as Vision Transformer (ViT)~\cite{dosovitskiy2020image} and SWIN Transformer~\cite{liu2021swin}, demonstrate the growing significance of transformers in image classification.

In the context of artwork classification, large-scale artwork datasets like Art500k~\cite{mao2017deepart}, OmniArt~\cite{strezoski2018omniart}, and the Rijksmuseum challenge dataset~\cite{Mensink2014TheRC} have provided benchmarks for classification in the artistic domain.

Fine-tuning pre-trained networks trained on large-scale natural image datasets like ImageNet \cite{russakovsky2015imagenet} has become a conventional method for analyzing artworks \cite{Gonthier2020AnAO,Sabatelli2018DeepTL,Zhao2022BigTL}. Cetinic et al.~\cite{Cetinic2018FinetuningCN} use CNNs for a series of art-related applications. Similarly, Hong et al.~\cite{Hong2017ArtPI} provide a method to differentiate artwork images from a variety of angles and perspectives. Different methods have been introduced to classify artists, genres, and material in the artwork datasets \cite{Agarwal2015GenreAS, Rodriguez2018ClassificationOS,Bar2014ClassificationOA,LiIeeeTO, Sablatnig1998HierarchicalCO}. Saleh et al.~\cite{saleh2016large} investigate a comprehensive list of visual features and metric learning approaches to learn an optimized similarity measure between paintings.

Key developments in gesture recognition within artwork images have primarily been driven by tailored datasets such as SniffyArt~\cite{zinnen2023sniffyart}, DEArt~\cite{reshetnikov2022deart} and PoPArt \cite{schneider2023poses} with a focus on pose-related gestures.

Our work aims to contribute to the evolving field of computational art history, specifically with a focus on uncommon senses, like smell. We adapt and extend real-world object detection and classification methods with the aim to automatically recognize smell gestures in historical artworks.

\section{Dataset and Challenges}

\begin{figure}[!htb]
    \centering
    \includegraphics[width=0.8\linewidth]{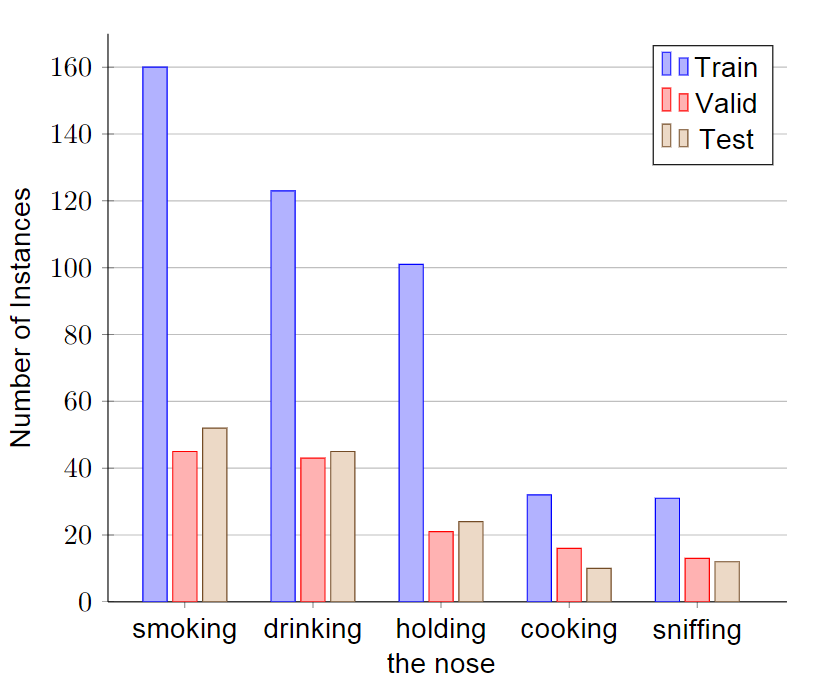}
    \caption{Class Distribution Excluding Background Class. It is important to note that the dataset has not only a small number of samples per class but also a significant class imbalance. \textit{Figure taken from~\cite{zinnen2023sniffyart} with permission to reuse granted by the authors.}}
    \label{fig:dista}
\end{figure}

The SniffyArt dataset comprises 1941 persons annotated with 6 gesture classes displayed in~\Cref{fig:classes_example}. 
Several challenges aggravate the training of algorithms to reliably recognize smell gestures and generalize to unseen data: 

\begin{enumerate}
    \item Compared to large-scale datasets, it has only very few training samples, which makes training deep-learning models difficult.
    \item The training (and test) samples exhibit a significant class imbalance. The~\Cref{fig:dista} illustrates the skewed data distribution across the training, validation, and test sets.
    \item A common method to compensate for small training data is incorporating external training data, either for pre-training or to enrich the dataset. 
    However, the process of gathering new gesture annotations is time-consuming and costly. Artwork images often lack detailed meta-data, specifically regarding olfactory dimensions of the artwork~\cite{ehrich2021nose}, and obtaining additional labeled data to address the class imbalance becomes a challenging task. 
\end{enumerate}

\section{Proposed Method}

\begin{figure}[!ht]
    \centering
    \includegraphics[width=0.6\linewidth]{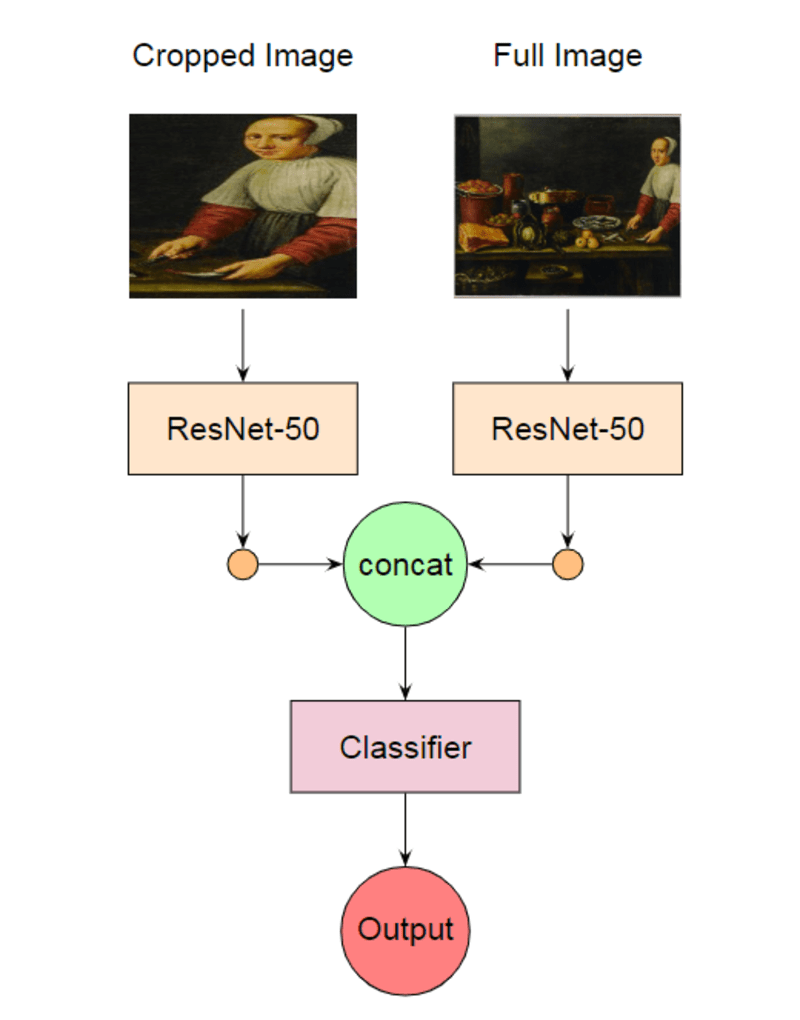}
    \caption{Architecture diagram for the proposed model. The cropped person and the full context image are passed through separate backbones. Finally, the outputs of these backbones are concatenated and passed to the classifier.}
    \label{fig:architecture}
\end{figure}

Similar to the work of Kosti et al.~\cite{kosti2017emotion}, our model architecture comprises three main modules: two feature extractors and a fusion module. The first module, dedicated to capturing detailed information about the person performing gestures, focuses on the region of the image corresponding to that person instance. This module efficiently extracts the most relevant features associated with the person's actions.

Simultaneously, the second module is designed to process the entire image, extracting global features that provide essential contextual support. By incorporating the entire image in this module, our model gains access to a wealth of information beyond the immediate region of the person of interest. This broader perspective enables the model to capture the relationships between the individual's actions and the environmental context in which they unfold. For instance, in cooking scenarios (~\Cref{fig:sub6}), understanding not only the specific movements of the chef but also the layout of the kitchen, the presence of utensils, and the arrangement of ingredients becomes crucial for accurate gesture interpretation.

Finally, the third module takes both the image and person features as input and performs the fusion and gesture classification using a four-layer fully connected neural network (FCNN). The parameters of all three modules are learned jointly. The overall architecture of our proposed method is illustrated in~\Cref{fig:architecture}. 

For doing inference on new unseen images, our approach requires the identification of individual persons. For this purpose, any object detection technique can be employed to provide the initial person detection. The performance of this pre-processing step can be optimized by fine-tuning the detection models with artwork datasets with annotated persons, e.\,g.\ PeopleArt~\cite{westlake2016detecting}, DEArt~\cite{reshetnikov2022deart}, and PoPArt~\cite{schneider2023poses}.
After the extraction of persons, both the cropped person and the entire image are fed into the respective branches of our classification network.


\section{Results}

When incorporating context information using our proposed method, we observe a notable improvement across all backbones, as shown in~\Cref{table:results_comparison}. This improvement suggests that the inclusion of contextual information is important for the model's learning capabilities. The contextual features extracted from the entire artwork scene augment the model's understanding of gestures within the broader artistic context.

Surprisingly, transformer-based methods, such as HRNet-W32~\cite{wang2020deep} and SwinV2~\cite{liu2022swinv2}, generally underperform compared to their ResNet~\cite{he2016deep} counterparts. This discrepancy may be attributed to the limited effectiveness of ImageNet pre-trained weights for the specific task of gesture recognition. To improve the performance of Transformer-based models, it is likely that a larger gesture recognition dataset or pretraining on tasks closely related to gesture recognition would be necessary.

\begin{table}[!ht]
\centering
\caption{Performance Comparison using Different Backbones. We can see an improvement in F\textsubscript{1} score with the inclusion of context on all the backbones by notable margins. }
\begin{adjustbox}{max width=\textwidth}
\begin{tabular}{llc}
\toprule
\textbf{Backbone} & \textbf{} & \textbf{Test-F1} \\
\midrule
\multirow{3}{*}{HRNet-W32 \cite{wang2020deep}}  
  & Without Context \cite{zinnen2023sniffyart} & 17.3 ± 1.8 \\
  & \textbf{With Context (ours)} & \textbf{37.1 ± 2.8} \\
\midrule
\multirow{3}{*}{ResNet-101 \cite{he2016deep}}  
  & Without Context \cite{zinnen2023sniffyart} & 34.2 ± 1.8 \\
  & \textbf{With Context (ours)} & \textbf{36.7 ± 1.9} \\
\midrule
\multirow{3}{*}{ResNet-50 \cite{he2016deep}}   
  & Without Context \cite{zinnen2023sniffyart} & 31.1 ± 2.0 \\
  & \textbf{With Context (ours)} & \textbf{36.8 ± 1.9} \\
\midrule
\multirow{3}{*}{SwinV2 \cite{liu2022swinv2}} 
  & Without Context & 15.8 ± 1.9 \\
  & \textbf{With Context (ours)} & \textbf{18.7 ± 1.9} \\
\bottomrule
\label{table:results_comparison}
\end{tabular}
\end{adjustbox}
\end{table}

\section{Conclusion and Future Directions}

In this paper, we demonstrate that our proposed approach, which considers both cropped and context images, improves the classification F\textsubscript{1} score compared to baseline methods with the same backbone. Specifically, it shows improvements of approximately 5\%, 3\%, and 20\% on ResNet-50, HRNet-W32, ResNet-101, and SwinV2, respectively.  

As our research progresses, we aim to incorporate multimodal learning by leveraging pose estimation keypoints using transformer-based backbones and fusion methods. This approach can help us enhance our classification performance even more due to their effectiveness proven by recent research \cite{carion2020end} \cite{zhang2022dino}, \cite{liu2021swin}.

Alternatively, we are considering a broader approach focused on extending the dataset with more general activities. By initially targeting a more general task, we aim to establish a foundation for understanding overall actions occurring in artworks. Image representations trained on these broader categories can potentially serve as a starting point to fine-tune for the more specific task of recognizing smell gestures. This approach will also help address the current issues with the SniffyArt dataset~\cite{zinnen2023sniffyart}.

At its current stage, this work presents itself as one step forward towards the aim of automatically recognizing smell gestures within past visual culture, eventually promoting the role of uncommon senses, such as smell, in digital humanities and computational heritage.

\section*{Acknowledgments}
This paper has received funding from the Odeuropa EU H2020 project under grant agreement No.\ 101004469. 
We gratefully acknowledge the donation of the NVIDIA corporation of two Quadro RTX 8000 that we used for the experiments.

\clearpage
\bibliographystyle{plain}

\end{document}